\documentclass[sn-aps]{sn-jnl}

\usepackage{graphicx}%
\usepackage{multirow}%
\usepackage{amsmath,amssymb,amsfonts}%
\usepackage{amsthm}%
\usepackage{mathrsfs}%
\usepackage[title]{appendix}%
\usepackage{xcolor}%
\usepackage{textcomp}%
\usepackage{manyfoot}%
\usepackage{booktabs}%
\usepackage{algorithm}%
\usepackage{algorithmicx}%
\usepackage{algpseudocode}%
\usepackage{listings}%
\usepackage{algpseudocode,algorithm}
\usepackage{soul}
\usepackage{dsfont}
\usepackage{tabularx}
\usepackage{fancyhdr}
\fancypagestyle{firstpage}{
  \fancyhf{} 
  \rfoot{} 
  \cfoot{} 
  \lfoot{This work was carried out while Alessandro was a Master's student and Davide was a Ph.D. student at DTU. Alessandro is now with Novo Nordisk A/S, and Davide is with EDF Trading.} 
}

\begin{document}

\title[Article Title]{Data-driven jet fuel demand forecasting: A case study of Copenhagen Airport}


\author{\fnm{Alesssandro} \sur{Contini}}

\author{\fnm{Davide} \sur{Cacciarelli}}

\author{\fnm{Murat} \sur{Kulahci}}\email{muku@dtu.dk} 

\affil{\orgdiv{Department of Applied Mathematics and Computer Science}, \orgname{Technical University of Denmark (DTU)}, \orgaddress{\city{Kgs. Lyngby}, \country{Denmark}}}


\abstract{Accurate forecasting of jet fuel demand is crucial for optimizing supply chain operations in the aviation market. Fuel distributors specifically require precise estimates to avoid inventory shortages or excesses. However, there is a lack of studies that analyze the jet fuel demand forecasting problem using machine learning models. Instead, many industry practitioners rely on deterministic or expertise-based models. In this research, we evaluate the performance of data-driven approaches using a substantial amount of data obtained from a major aviation fuel distributor in the Danish market. Our analysis compares the predictive capabilities of traditional time series models, Prophet, LSTM sequence-to-sequence neural networks, and hybrid models. A key challenge in developing these models is the required forecasting horizon, as fuel demand needs to be predicted for the next 30 days to optimize sourcing strategies. To ensure the reliability of the data-driven approaches and provide valuable insights to practitioners, we analyze three different datasets. The primary objective of this study is to present a comprehensive case study on jet fuel demand forecasting, demonstrating the advantages of employing data-driven models and highlighting the impact of incorporating additional variables in the predictive models.}





\maketitle
\thispagestyle{firstpage} 

\section{Introduction}
\label{sec:intro}
Fuel consumption is a critical area of research in the transportation business and management. Within the aviation industry, jet fuel represents one of the largest operational expenses for airline companies \cite{Morrell}. To mitigate the risks associated with sudden price fluctuations, many airlines employ fuel hedging strategies \cite{LIM201433}. A commonly used hedging strategy is the jet fuel swap, where a volatile-priced commodity is exchanged at a fixed price over a specified period. Airlines optimize this strategy by considering their tentative flight schedules, which are typically publicly available for a long-term horizon. Fuel providers also rely on these schedules to inform their decision-making processes regarding fuel sourcing. However, discrepancies often arise between the scheduled and actual number of flights, leading to the potential for under- or over-stocking. Accurate estimation of fuel demand is thus crucial for achieving an efficient sourcing strategy. Despite numerous studies focusing on macroeconomic fuel demand forecasting, there is a paucity of case studies that address short-term jet fuel demand forecasting from a supply chain perspective. Furthermore, many industry practitioners still rely on basic models and primarily depend on tentative schedules provided by airlines.

This study aims to evaluate the effectiveness of machine learning models in accurately predicting jet fuel demand in the Danish market, utilizing data collected from the largest aviation fuel distributor in the country. The project encompasses three distinct case studies. The first two cases investigate the fuel purchasing patterns of a flag carrier and a low-cost carrier, respectively. The third case study focuses on the total fuel volume ordered by all airlines operating through Copenhagen Airport (CPH). The analysis includes only airlines that are existing customers of the studied fuel distributor, which still represents a substantial portion of the CPH market, serving approximately 35\% of airlines flying through CPH (as of August 2022).

The primary challenge in this research lies in ensuring the availability of high-quality data. Reliable predictions necessitate a significant historical dataset that is both extensive and meaningful. However, obtaining such data can be challenging. Additionally, there are common issues in this field related to the presence of unpredictable factors that can significantly impact demand or supply. For instance, the recent COVID-19 pandemic resulted in reduced flights due to restrictions, and the sudden acquisition of new contracts and routes by airlines can profoundly affect flight volumes. Inadequate planning and stocking based on incomplete or inaccurate data can result in substantial losses, even for large and well-organized companies. To ensure success and resilience in the face of such changes, it is crucial to prioritize data quality and address any limitations or biases present in the dataset.

The remainder of this paper is organized as follows. Section \ref{sec:related} examines existing literature and related problems, providing motivation for the chosen methodologies. In Section \ref{sec:meth}, a summary of the machine learning models employed in the experiments is presented. Section \ref{sec:exp} evaluates the effectiveness of the data-driven strategies through the analysis of the three case studies. Finally, Section \ref{sec:end} provides some conclusions.

\section{Related work}
\label{sec:related}

Accurately forecasting jet fuel demand has significant economic and managerial implications. However, the current research on data-driven approaches to address this task is limited, particularly when it comes to exploring the potential of deep learning or time series models for forecasting aviation fuel demand. Existing studies in this domain have primarily focused on modeling the consumption of individual aircraft rather than market demands. For example, Li et al. \cite{consumption1} and Bauman et al. \cite{consumption2} conducted studies on forecasting consumption based on aircraft and flight-specific statistics, such as wind and time, while considering aircraft type. Another area of research in this field is flight time prediction, where techniques such as those used by Zhu et al. \cite{loadings} used deep learning methods to predict flight time for more efficient fuel loading scheduling. Although these studies are relevant, our research aims to take a broader perspective by assessing jet fuel demands using a purely data-driven approach.

Despite the limited studies specifically focused on jet fuel forecasting, demand forecasting can effectively be addressed through time series analysis, which employs various techniques to predict future values based on historical data. These techniques range from simple regression and basic time series models to more advanced methodologies. In the energy domain, researchers have used a plethora of methods to forecast future consumption \cite{regrex}. The two primary classes of methods that have been widely used for demand forecasting are traditional statistical methods and deep learning methods. Among the statistical methods, seasonal autoregressive integrated moving average (SARIMA) models have historically been successful \cite{SARIMAX_approach, sarimax_ANN_Natual_Gas2, chan2019}. Another statistical model that has gained popularity for time series forecasting is Prophet, developed by Facebook, which can handle diverse time series patterns including seasonality and trends \cite{prophet1,prophet2,prophet,prophet4}. Deep learning-based methods, particularly artificial neural networks (ANNs) and long short-term memory (LSTM) networks, have gained increasing popularity in recent years for time series problems \cite{KARASU2022122964,8859013,9336914}. Finally, hybrid methods that combine statistical models with deep learning-based approaches have also been proposed \cite{arima+lstm_hybrid,sarima_lstm_hybrid,natural_gas_daily_SARIMAX_LSTM_hybr,Javad}. The use of hybrid models aims to efficiently capture both linear and non-linear components of the data \cite{SULANDARI2020116408}, combining the interpretability and simplicity of linear models \cite{SBAL,ROAL,cacciarelli2024real,cacciarelli2023stream} with the accuracy of deep learning models. Examples of semi-supervised learning combining linear models with deep learning models such as autoencoders are presented in \cite{cacciarelli2025semi,cacciarelli2022online}.

In summary, although there is a lack of research on data-driven approaches for forecasting jet fuel demand, there is potential to apply advanced  machine learning techniques to improve the accuracy and reliability of predictions and optimize the sourcing operatians. By leveraging these approaches and considering the broader perspective of market demands, our research aims to contribute to the advancement of jet fuel demand forecasting.

\section{Methods}

\label{sec:meth}
\subsection{SARIMA}
The SARIMA model is a popular class of time series models that extend the autoregressive moving average (ARMA) models for dealing with non-stationary and seasonal time series. At its core, the autoregressive (AR) component relates the current observation $y_t$ of a time series to its previous observations, $y_{t-1}, y_{t-2}, \ldots, y_{t-p}$, in a similar way to a regression model \cite{murat1,murat2}. In general, an AR model of order $p$, $\text{AR}(p)$, is given by
\begin{equation}
    y_t = \phi_1 y_{t-1} +\phi_2 y_{t-2} + \cdots + \phi_p y_{t-p} + a_t
\end{equation}

\noindent
where $a_t$ is an error term assumed to be uncorrelated with mean $E(a_t)=0$ and constant variance $Var(a_t)=\sigma_a^2$. The regression coefficients $\phi_i$ are parameters that need to be estimated from the data. The second key component of an ARMA model is the moving average (MA) term, which is used to average past and present noise terms. An MA model of order $q$, $\text{MA}(q)$, is given by
\begin{equation}
    y_t = a_t - \theta_1 a_{t-1} -\theta_2 a_{t-2} - \cdots - \theta_q a_{t-q} 
\end{equation}

\noindent
where the coefficients $\theta_i$ are estimated from the data. While in pure AR models the current observation is estimated using the lagged values of the same time series, in pure MA models the prediction takes into account the errors/residuals of the previous forecasts. The errors might be considered innovations, novelties that the MA model failed to catch. So, while in AR models we average across recent observations, in MA models we average across recent innovations. The combined ARMA$(p,q)$ model can be written as
\begin{equation}
    y_t = \phi_1 y_{t-1} +\phi_2 y_{t-2} + \cdots + \phi_p y_{t-p} + a_t - \theta_1 a_{t-1} -\theta_2 a_{t-2} - \cdots - \theta_q a_{t-q} 
\end{equation}

\noindent
Standard ARMA models only apply to stationary time series, which are in equilibrium around a constant mean and are characterized by a distribution that is independent of time shifts \cite{murat1}. From an applied point of view, non-stationarity can often be visually identified by the presence of large swings up and down in the time series. However, there are more rigorous statistical tests that can be performed to verify if a time series is stationary \cite{priestley1969test,witt1998testing}. The simplest level of non-stationarity is often dealt with using a first-order differencing, which is defined as $\nabla y_t = y_t - y_{t-1}$. Because the inverse operation of differencing is summing or integrating, an ARMA$(p,q)$ model applied to a $d$ differenced data is defined as an ARIMA$(p,d,q)$ model. When the data is also affected by seasonal behavior, we might also use a seasonal difference operator like $\nabla_s y_t = y_t - y_{t-s}$, where $s$ identifies the length of one seasonal period (e.g., $s=7$ for a weekly pattern). A seasonal behavior is typically characterized by the presence of repeating and cyclic patterns. Once a seasonal component is deemed necessary to model the time series of interest, SARIMA models can be employed to incorporate both the non-seasonal and seasonal components into a model, effectively capturing the complex dynamics of a non-stationary and seasonally dependent time series. The complete model is defined as a SARIMA$(p, d, q) \times (P, D, Q, s)$ model, where $p,d,q$ are the orders of the non-seasonal component and $P,D,Q,s$ the ones for the seasonal component. The seasonal parameters, $P,D,Q$, are similar to their non-seasonal counterparts but operate on the seasonal differences of the data. 
Finally, SARIMA models with exogenous variables (SARIMAX) can be employed to enhance the predictive performance of the models. The inclusion of exogenous variables allows for better handling of sudden changes in levels that cannot be identified solely based on previous observations. When including exogenous variables in a SARIMAX model, the estimation process involves jointly estimating the parameters of the autoregressive, differencing, and moving average components, as well as the coefficients associated with the exogenous variables.

\subsection{Prophet}
Prophet is a time series model developed by Facebook \cite{prophet}, which is based on the decomposable time series model previously proposed by Harvey and Peters \cite{decomposable}. The model is defined by

\begin{equation}
    y(t) = g(t) + s(t) + h(t) + e(t)
\end{equation}

\noindent
where $g(t)$ represents the trend component, capturing non-periodic changes over time and modeled as a piecewise linear or logistic function, $s(t)$ represents the seasonal component, capturing periodic changes in the data, $h(t)$ represents the effects of holidays or other user-specified events that can be modeled as a binary indicator function, and $e(t)$ represents the error term or residual, accounting for any remaining variation not explained by the model components, which is assumed to be normally distributed. The Prophet model uses a Bayesian framework based on Markov Chain Monte Carlo (MCMC) sampling to estimate the model parameters. There are two key tunable parameters that can be adjusted to customize the forecasting behavior. The first one is the change point prior scale parameter, which controls the flexibility in detecting and modeling trend change points in the model. It determines the penalty applied to the rate of trend changes, affecting the number and timing of change points identified by the model.  A higher value makes the model more flexible in detecting trend changes, allowing for more potential change points. This can result in a model that captures more short-term fluctuations in the data. Conversely, a lower value imposes a stronger penalty on the number of change points, leading to a smoother trend with fewer identified change points. This is useful when there is a need to emphasize long-term trends and avoid overfitting to short-term fluctuations. Another important hyperparameter is the seasonality prior scale, which controls the flexibility of the seasonal component in the model. It represents the strength of the prior imposed on the seasonal component during the model fitting process. A high value allows the model to be more flexible and adapt to the seasonal patterns in the data more closely while a lower value results in a more rigid model, which may be useful when the observed seasonality is less pronounced or when there is a need to downplay the impact of seasonality on the forecast.

\subsection{LSTM}
\begin{figure}
\caption{An LSTM cell and its components: $h_t$ represents the hidden state, $c_t$ is the cell state and $x_t$ is the input value at time $t$.}
\label{fig:lstmCell}
\centering
\includegraphics[width=.6\textwidth]{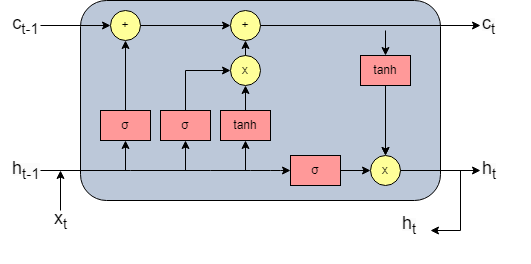}
\end{figure}

Recurrent neural networks (RNNs) are usually preferred to traditional methods when the time series is particularly complex. The complexity may be due to nonlinearities or to the fact that trends and seasonal patterns change over time. In an RNN network, there are loops that make the current output depend on the previous hidden states. An LSTM neural network is a type of RNN specifically designed to capture long-term dependencies in sequential data \cite{LSTM_architect_references}. LSTM networks introduce memory cells and gating mechanisms that enable the network to selectively remember or forget information over multiple time steps, thus alleviating the vanishing gradient problem and allowing for the modeling of long-term dependencies. The connection between LSTM and ARIMA models is discussed in \cite{cacciarelli2025modeling}. The structure of an LSTM cell is shown in Figure \ref{fig:lstmCell}. 
The cell state $c_t$ can be considered the long-term memory, and the hidden state $h_t$ as the short-term memory. There are three types of gates that control the flow of information through the network. The forget gate $f_t$ determines which information from the previous memory cell state $c_{t-1}$ should be discarded or forgotten. It outputs a value between 0 and 1 for each memory cell element, indicating the degree to which the corresponding element should be forgotten. The input gate $i_t$ is used to decide which new information should be stored in the memory cell. It generates a value between 0 and 1 for each memory cell element, representing the extent to which the corresponding element should be updated. The output gate $o_t$ determines how much information from the memory cell should be exposed or outputted. It produces a value between 0 and 1 for each memory cell element, indicating the extent to which the corresponding element should influence the current hidden state (h(t)). The forget, input, and output gates take takes as input the previous hidden state $h_{t-1}$ and the current input $x_t$. The current cell state $c_t$ is updated based on the forget gate, input gate, and new input as in

\begin{equation}
    c_t = f_t \cdot c_{t-1} + i_t \cdot g_t
\end{equation}

\noindent
where $f_t$ is the forget gate output, $c_{t-1}$ is the previous memory cell state, $i_t$ is the input gate output, and $g_t$ is the new candidate cell state. The hidden state $h_t$ is computed based on the output gate and the updated memory cell

\begin{equation}
    h_t = o_t \cdot tanh(c_t)
\end{equation}

\noindent
where $o_t$ is the output gate output and $tanh$ is the hyperbolic tangent function. These update equations allow the LSTM network to control the flow of information by selectively remembering or forgetting information at each time step. LSTM networks are trained using backpropagation through time, where the gradients are calculated through time and used to update the model parameters. In this study, the LSTM neural networks have been trained using a loss function known as negative log-likelihood (NLL), which is defined as

\begin{equation}
\text{NLL} = - \sum_{i=1}^n\log(p(y_i | f(x_i; \theta), \sigma))
\end{equation}

\noindent
where $f(x_i;\theta)$ represents the output of the LSTM network for the $i$th sample, which is determined by the input $x_i$ and the model parameters $\theta$. The term $y_i$ refers to the $i$th observation of the dependent variable and $\sigma$ corresponds to the standard deviation of a normal distribution, which is a parameter that can be adjusted during the training process. The probability density function $p(y_i | f(x_i; \theta))$ represents the likelihood of the normal distribution evaluated at $y_i$, with a mean of $f(x_i;\theta)$ and a standard deviation of $\sigma$. Lastly, $n$ denotes the total number of observations. Using this particular loss function offers several advantages. By incorporating the trainable parameter $\sigma$, it becomes feasible to compute confidence intervals rapidly, enabling the estimation of overall uncertainty associated with the model's predictions.

\subsection{Hybrid models}
\label{subsec:hybrid}

\begin{figure}[h]
\caption{Schema of the SARIMAX-LSTM hybrid model.}
\label{fig:hybrid_schema}
\centering
\includegraphics[width=.8\textwidth]{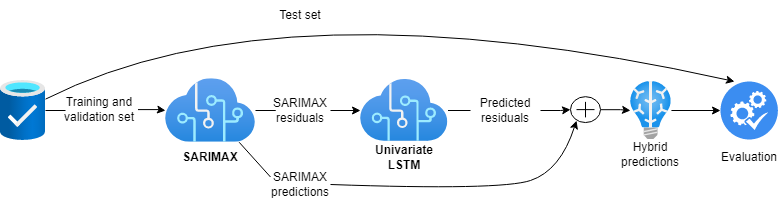}
\end{figure}

Hybrid models are created by combining multiple models to leverage their individual strengths. In this study, we examine the effectiveness of a hybrid architecture that merges statistical time series models with deep learning models. In particular, we examine the performance of an ensemble obtained by stacking a SARIMAX model and an LSTM neural network \cite{iterative_residuals_forecasting, SULANDARI2020116408}. The architecture of the model is illustrated in Figure \ref{fig:hybrid_schema}. The ensemble is designed to be relatively straightforward, employing a two-step forecasting process. Initially, a SARIMAX model is fitted to the original time series, followed by an LSTM network that explores the presence of any remaining patterns present in the residuals. The final prediction is obtained by combining the predictions of the two models. It is important to note that the performance of the hybrid architecture will only improve the original time series model if the residuals do not solely consist of unpredictable noise. In cases where the time series model accurately captures the data, extracting meaningful information from the residuals may prove challenging. Indeed, if the residuals of the first model are pure noise, without any learnable signal for the second model, combining the two predictions may actually degrade the overall performance.

\section{Case studies}
\label{sec:exp}

\subsection{Data collection}
The primary objective of this study is to provide the fuel provider with an advanced decision support tool for forecasting the fuel quantity purchased by each airline in the upcoming month. This tool aims to optimize the sourcing strategy and price negotiations. The data is automatically collected during fueling operations at the airport and stored in the company database. The main challenge of this study lies in the need for long-term forecasting, as the company requires estimates of fuel purchases for the entire upcoming month. Since the available data consists of daily records from 2019 to 2022, there was not a sufficient number of data points to develop predictive models with a monthly step. Consequently, models with a daily step have been used to predict the next 30 days. To explore the impact of time series patterns, data quality, and airline-specific differences, we focused on three distinct case studies. The first case study examines the consumption data of a major flag carrier, the second case study focuses on a low-cost carrier, and the third case study analyzes overall volumes of CPH. The target variable being predicted is the actual amount of fuel purchased by the airlines each day. Among the exogenous variables considered in this study, the most important one is the number of daily scheduled flights, which indicates the number of flights scheduled for a given day. However, it is important to note that this measure does not always correspond to the actual number of operated flights, as the prediction is based on the schedule provided by the airlines in the previous months for the upcoming month.

\subsection{Model selection and evaluation}
For all three case studies, the data was partitioned into training, validation, and test sets. The hyperparameters of the model were determined through a grid search optimization procedure, aiming to minimize the mean squared error (MSE) on the validation set. In the SARIMA and SARIMAX models, the procedure assisted in identifying the appropriate orders ($p, d, q$ and $P, D, Q$) for the models. The seasonal period $s$ was fixed at seven, considering the strong weekly seasonal pattern observed in the data. In the Prophet model, the grid search aided in determining the change point prior scale and seasonality prior scale parameters. As for the LSTM models, the tuning procedure was employed to determine the window size, number of LSTM layers, and hidden dimensionality. Notably, the window size plays a crucial role as it determines the number of input time steps used by the model to generate predictions for the next 30 days. The LSTM models were trained using the Adam optimizer with a learning rate of 0.01. Rather than setting a fixed number of epochs, an early stopping criterion was utilized, terminating training after 10 epochs without any observed improvement.

Regarding model evaluation, we employed 90 days of external test data to assess the prediction performance over three consecutive sourcing periods. The evaluation followed a walk-forward validation scheme \cite{Cacciarelli2021}. In this scheme, predictions were initially made for the next 30 days, followed by the observation of actual values, which were then used to update the model for predicting the subsequent 30 days. Root mean squared error (RMSE), mean absolute error (MAE), and symmetric mean absolute percentage error (SMAPE) were used as evaluation metrics. RMSE provides a comprehensive measure of prediction accuracy by quantifying the average magnitude of residuals between the predicted and actual values as in

\begin{equation}
\text{RMSE} = \sqrt{\frac{1}{n}\sum_{i=1}^{n}(y_i - \widehat{y}_i)^2}
\end{equation}

\noindent
where $y_i$ represents the actual value, $\widehat{y}_i$ represents the predicted value for the $i$th observation, and $n$ is the total number of observations. MAE, on the other hand, offers a simpler approach by calculating the average magnitude of residuals without squaring them, as in 

\begin{equation}
\text{MAE} = \frac{1}{n}\sum_{i=1}^{n}|y_i - \widehat{y}_i|
\end{equation}

\noindent
This metric focuses solely on the magnitude of errors and offers a more intuitive interpretation. SMAPE, another commonly used evaluation metric for time series forecasting, measures the percentage difference between the predicted and actual values, accounting for error magnitude relative to the data scale. It provides a symmetric assessment of percentage error, treating overestimation and underestimation errors equally. It is given by

\begin{equation}
\text{SMAPE} = \frac{1}{n}\sum_{i=1}^{n}\frac{|y_i - \widehat{y}_i|}{(|y_i| + |\widehat{y}_i|)/2} \times 100
\end{equation}

\subsection{Case study 1: flag carrier}
\label{sec:emirates}
\begin{figure}[h]
\caption{Scaled time series for the first case study, reporting the actual amount of fuel bought, the flight schedule of the same carrier that was available one month before, and the overall number of flights expected to depart from CPH airport.}
\label{img:emiratesVol}
\centering
\includegraphics[width=1\textwidth]{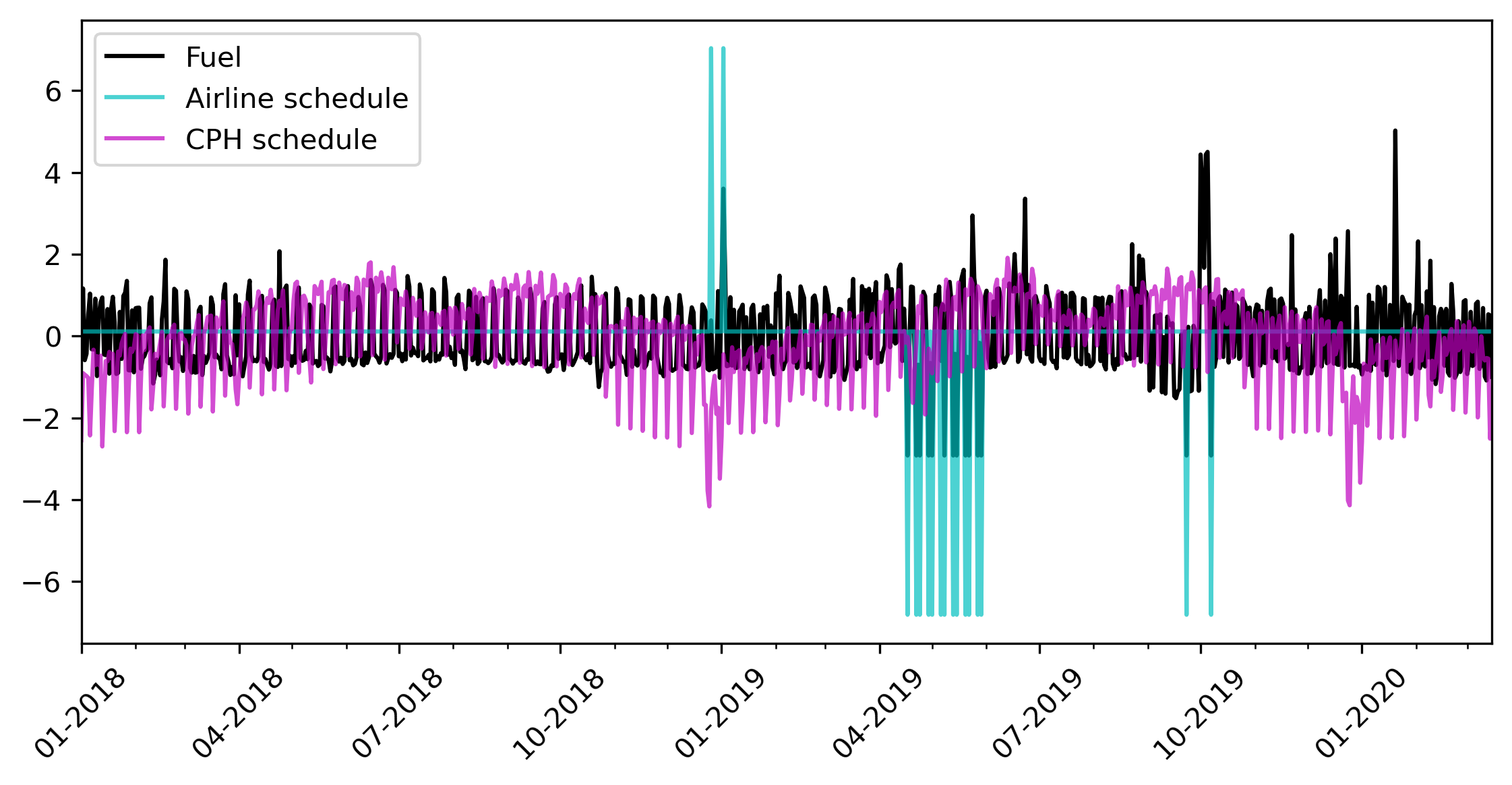}
\end{figure}

The first case study focuses on analyzing the fuel demand data of a large flag carrier, which mostly operates a small number of long-haul flights. The data provided by the company for this study spans from January 2018 to March 2020. Figure \ref{img:emiratesVol} shows that the fuel demand exhibits remarkable stability, with no significant volume fluctuations observed over the two-year period. The primary challenge in this case study stems from the fact that the tentative schedule available in the database does not reliably indicate the actual flights that will be operated by the carrier. This inconsistency in data could be attributed to data inaccuracies or the fact that the schedule may have been made available with less than one month notice. Additionally, it is worth noting that the other exogenous variable, which represents the overall number of flights scheduled to depart from CPH, does not exhibit a strong correlation with the actual amount of fuel purchased by the carrier.

\begin{figure}[h]
\caption{Test set predictions obtained with the SARIMA model on the first case study. The dashed red line represents the mean prediction and the shaded region corresponds to the 5\% confidence interval.}
\label{img:emi sarima}
\centering
\includegraphics[width=1\textwidth]{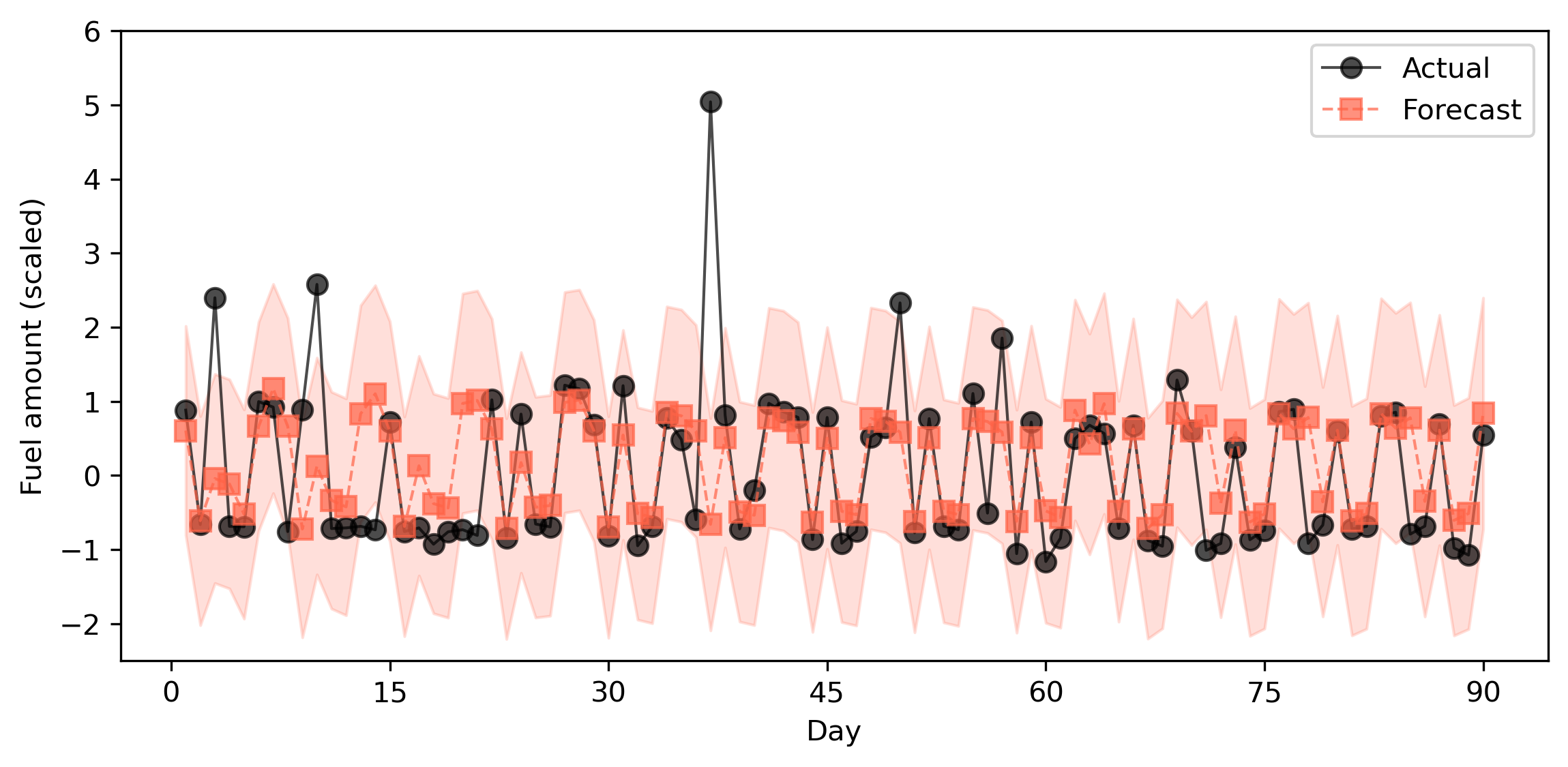}
\end{figure}

\begin{figure}[h]
\caption{Predicting the test set residuals of the SARIMA model with a univariate LSTM model on the first case study. The dashed orange line represents the mean prediction and the shaded region corresponds to the 5\% confidence interval.}
\label{img:emi resid}
\centering
\includegraphics[width=1\textwidth]{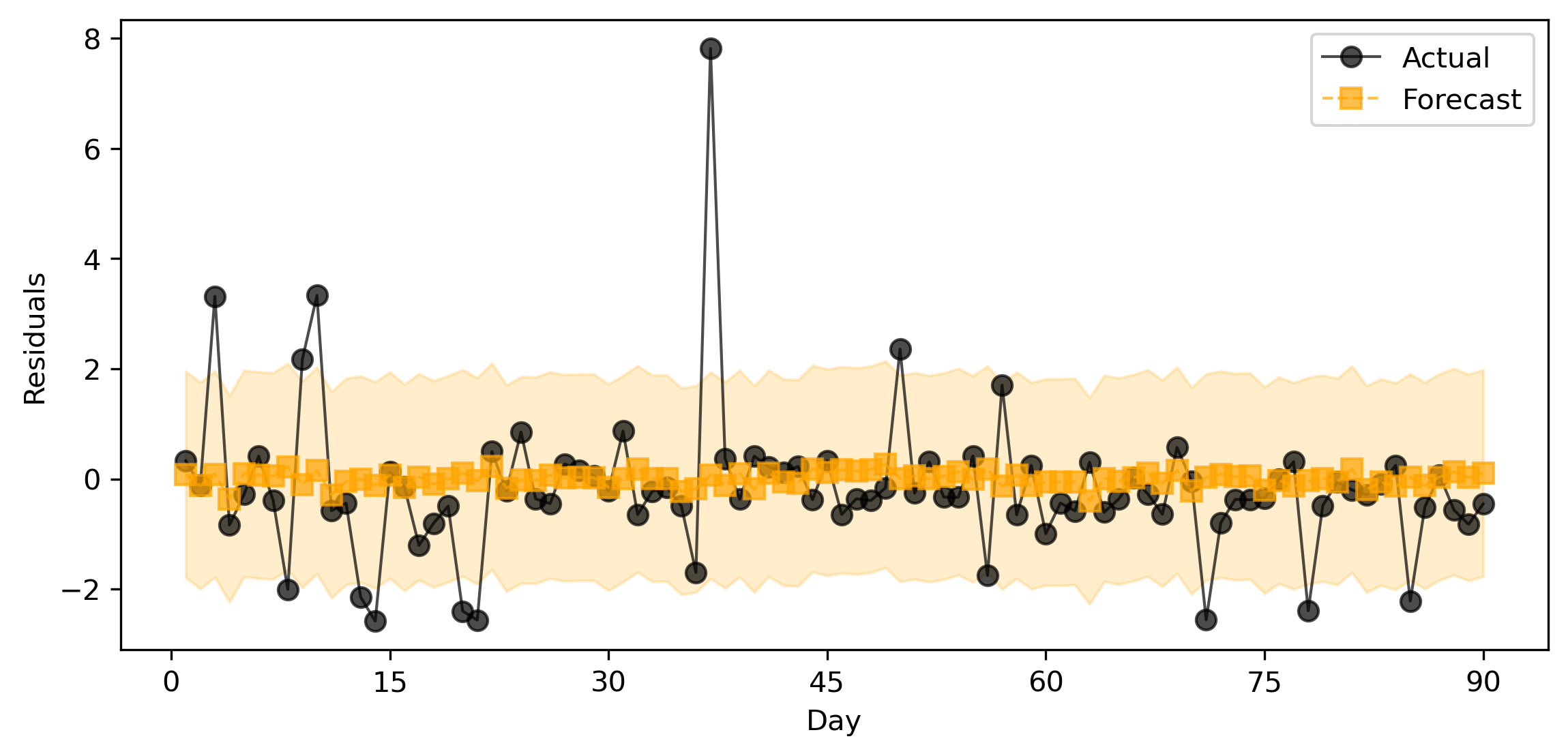}
\end{figure}

The results presented in Table \ref{tab:emirates_scores} indicate that the SARIMA model achieves the best predictive performance. The optimal parameters obtained through the grid search optimization procedure for the non-seasonal component are $(1,1,2)$, while for the seasonal component, the parameters are $(3,1,1,7)$. The predictions generated by this model on the test set are displayed in Figure \ref{img:emi sarima}. Although the SARIMA model may miss a few demand peaks, it accurately captures the overall behavior of the data. Interestingly, incorporating exogenous variables does not yield an improvement in performance across all models considered. This outcome was anticipated, as the reliability of the airline schedule is limited, and the other exogenous variable exhibits a weak correlation with the amount of fuel purchased. Moreover, the hybrid architecture does not enhance the prediction of the SARIMA model. Figure \ref{img:emi resid} demonstrates how the LSTM model accurately predicts only a few data points within the residuals. This observation suggests that the residuals mainly consist of unpredictable noise, with minimal remaining signal.

\begin{table} [h]
\centering
\begin{tabular}{@{}lccc@{}}
\toprule
\textbf{Model} & \textbf{RMSE} & \textbf{MAE} & \textbf{SMAPE} \\ 
\midrule
SARIMA & \textbf{0.971} & \textbf{0.569} & \textbf{16.985} \\
SARIMAX & 1.180 & 0.867 & 29.239 \\
Prophet univariate & 0.992 & 0.671 & 23.246 \\
Prophet multivariate & 0.994 & 0.677 & 23.567 \\
LSTM univariate & 1.030 & 0.644 & 21.970\\
LSTM multivariate & 1.083 & 0.684 & 22.001 \\
Hybrid (SARIMA-LSTM) & 0.975 & 0.583 & 17.652 \\ 
\bottomrule
\end{tabular}
\caption{Test set prediction performance of the different methods on the first case study.}
\label{tab:emirates_scores}
\centering
\end{table}

\subsection{Case study 2: low-cost carrier}
\label{sec:Low-cost carrier}

\begin{figure}
\caption{Scaled time series for the second case study, reporting the actual amount of fuel bought, the flight schedule of the same carrier that was available one month before, and the overall number of flights expected to depart from CPH airport.}
\label{img:ryVolumes}
\centering
\includegraphics[width=1\textwidth]{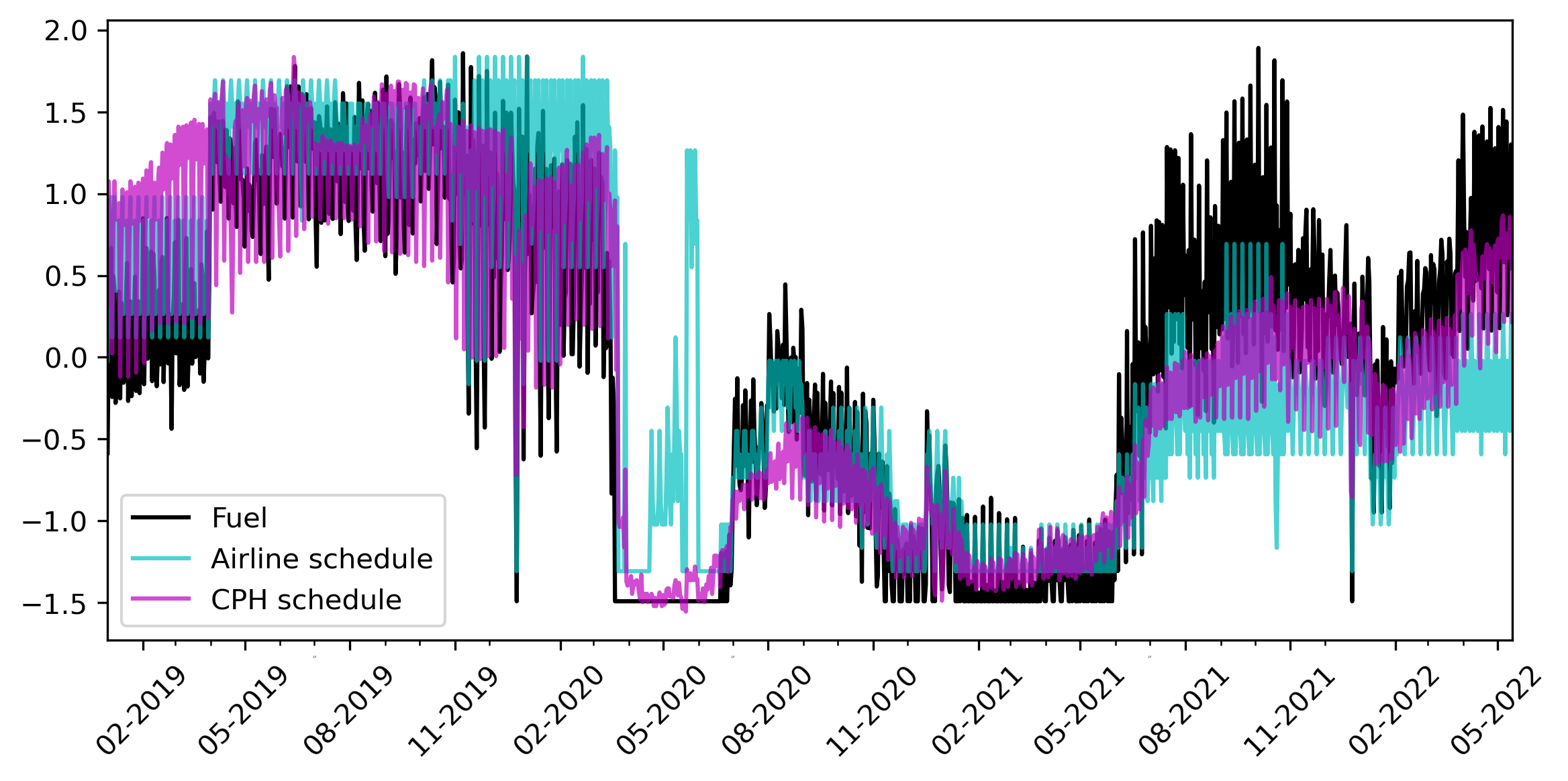}
\end{figure}

The data available for the second case study, as shown in Figure \ref{img:ryVolumes}, partially overlaps with the data used in the first case study. Higher volatility is observed in the ordered volumes for this case study. This increased volatility can be attributed to the carrier's operational characteristics, which involve a higher number of short-haul flights and a different sourcing strategy. The significant fluctuations depicted in the plot are also influenced by the temporary suspension of flights during the peak of the COVID-19 pandemic. Importantly, in this case study, the exogenous variables exhibit a strong correlation with the amount of fuel ordered by the company. Specifically, the airline schedule serves as a reliable indicator of future activity, with the exception of the months affected by the COVID-19 outbreak. The company demonstrates a high level of accuracy in publishing and updating its flight schedule. Additionally, the frequent small-volume fuel purchases made by the airline help to mitigate any disparities between scheduled and actual flight times.

\begin{figure}
\caption{Test set predictions obtained with the univariate LSTM model on the second case study. The dashed blue line represents the mean prediction and the shaded region corresponds to the 5\% confidence interval.}
\label{img:emi lstm}
\centering
\includegraphics[width=1\textwidth]{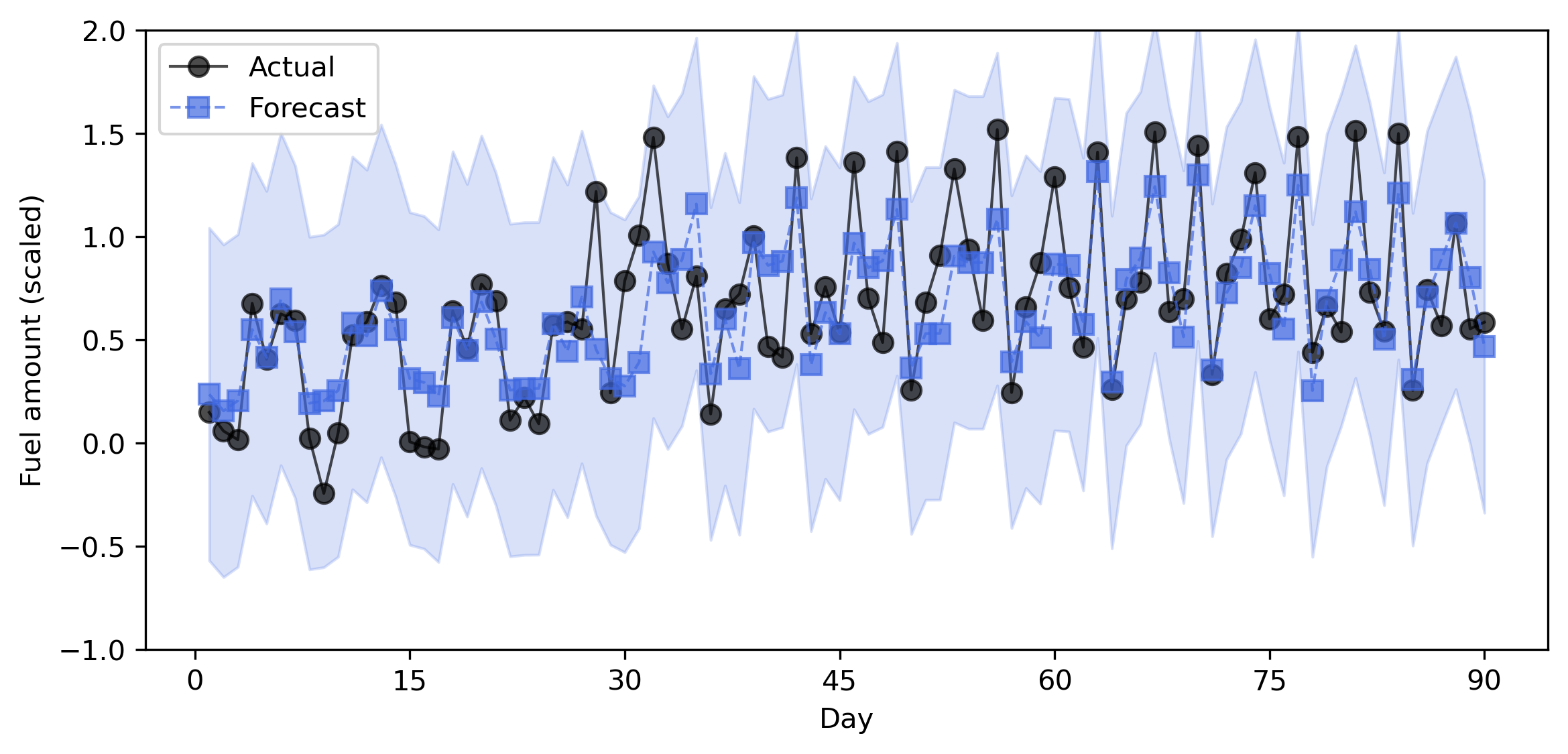}
\end{figure}

\begin{figure}
\caption{Test set predictions obtained with the SARIMA model on the second case study. The dashed red line represents the mean prediction and the shaded region corresponds to the 5\% confidence interval.}
\label{img:ry sarima}
\centering
\includegraphics[width=1\textwidth]{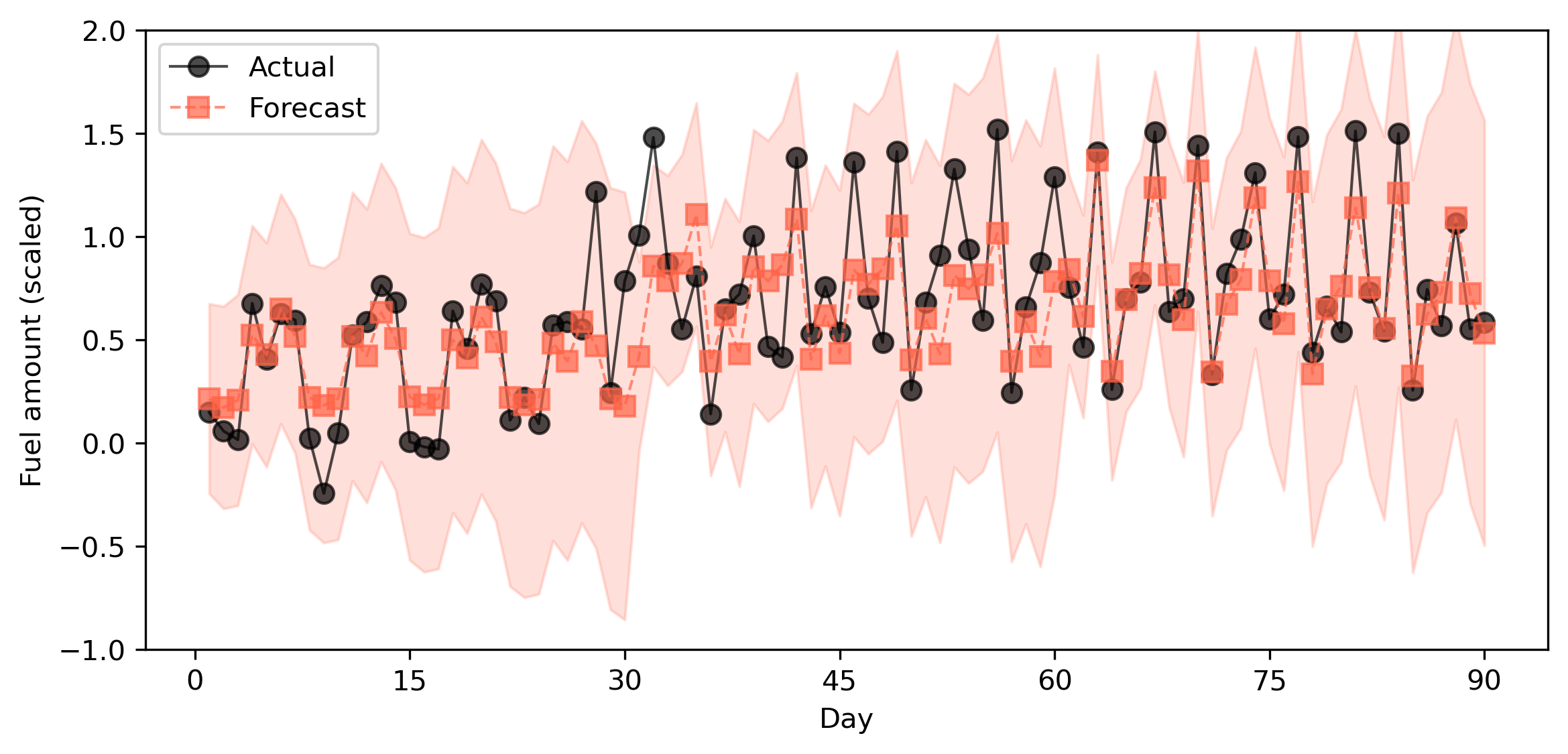}
\end{figure}

The prediction performances, as reported in Table \ref{tab:ryanair_scores}, show how the best forecasting model is now represented by the univariate LSTM network. However, the SARIMA model is extremely close in terms of predictive ability. The predicted values obtained by the LSTM model on the test set are shown in Figure \ref{img:emi lstm}. We can see how the highest forecasting errors appear at the beginning of the second month, when the volumes significantly increase from the previous 30 days. However, after a short time frame, the model is able to adapt and adeptly capture the underlying data patterns. A similar behavior is observed in the predictions of the SARIMA model, which are shown in Figure \ref{img:ry sarima}. In this case, the optimal order of the SARIMA model is given by $(1,0,2)\times(1,0,1,7)$. With regards to the Prophet model, in this case the multivariate version is able to outperform its univariate counterpart, highlighting the enhanced usefulness of incorporating exogenous variables compared to the first case study. With regard to the hybrid model, the LSTM network fitted on the residuals of the SARIMA model is not able to extract any meaningful pattern and the stacked model performs worse than the two individual models.

\begin{table}
\centering
\begin{tabular}{@{}lccc@{}}
\toprule
\textbf{Model} & \textbf{RMSE} & \textbf{MAE} & \textbf{SMAPE} \\ 
\midrule
SARIMA  & 0.253 & 0.195 &  10.359 \\ 
SARIMAX  & 0.329 & 0.247 & 12.641 \\ 
Prophet univariate & 0.462 & 0.384 & 21.529 \\
Prophet multivariate & 0.411 & 0.349 & 19.655 \\
LSTM univariate  & \textbf{0.248} & \textbf{0.194} & \textbf{10.230} \\ 
LSTM multivariate  & 0.261 & 0.220 & 12.350 \\ 
Hybrid (SARIMA-LSTM)  & 0.443 & 0.352 & 17.555 \\
\bottomrule
\end{tabular}
\caption{Test set prediction performance of the different methods on the second case study.}
\label{tab:ryanair_scores}
\centering
\end{table}

\subsection{Case study 3: CPH airport}
\label{sec:cph}
\begin{figure}
\caption{Scaled time series for the third case study, reporting the actual amount of fuel bought, the flight schedule of the same carrier that was available one month before, and the overall number of flights expected to depart from CPH airport.}
\label{img:cphVolumes}
\centering
\includegraphics[width=1\textwidth]{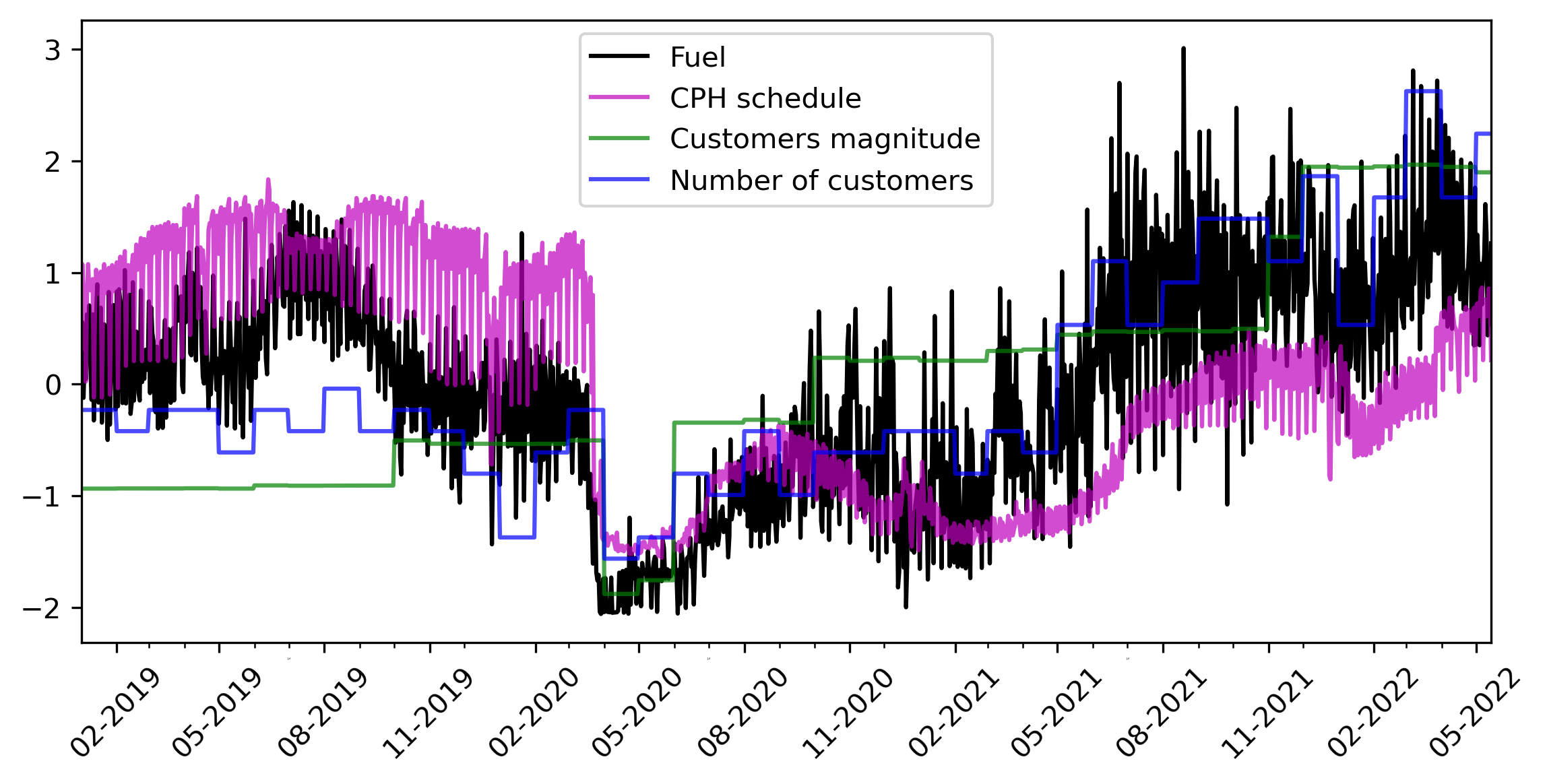}
\end{figure}

CPH Airport, situated in Kastrup, Denmark, is one of the largest airports in northern Europe. It serves not only the Copenhagen area but also a wide geographical area encompassing the rest of Zealand, the Øresund Region, and a significant portion of southern Sweden, including Scania. The total volume of fuel traded at CPH is influenced by a diverse customer base, each exhibiting distinct fueling behavior. These customers vary significantly in terms of fleet size, frequency of fueling, individual volume requirements, seasonality patterns, and operational reliability. Consequently, it is plausible that smaller companies with less predictable behavior may be overshadowed in predictions by larger, more consistently operating companies. The time series reported in Figure \ref{img:cphVolumes} spans the same period used in the second case study. The trend of the data is quite similar to the one reported in Figure \ref{img:ryVolumes}, and the key difference is the use of different exogenous variables. Indeed, in this case, we tried to add more information to the overall CPH schedule by detailing information about the number and magnitude of customers operating their flights.

\begin{figure}
\caption{Test set predictions obtained with the SARIMAX model on the third case study. The dashed red line represents the mean prediction and the shaded region corresponds to the 5\% confidence interval.}
\label{img:cph sarima}
\centering
\includegraphics[width=1\textwidth]{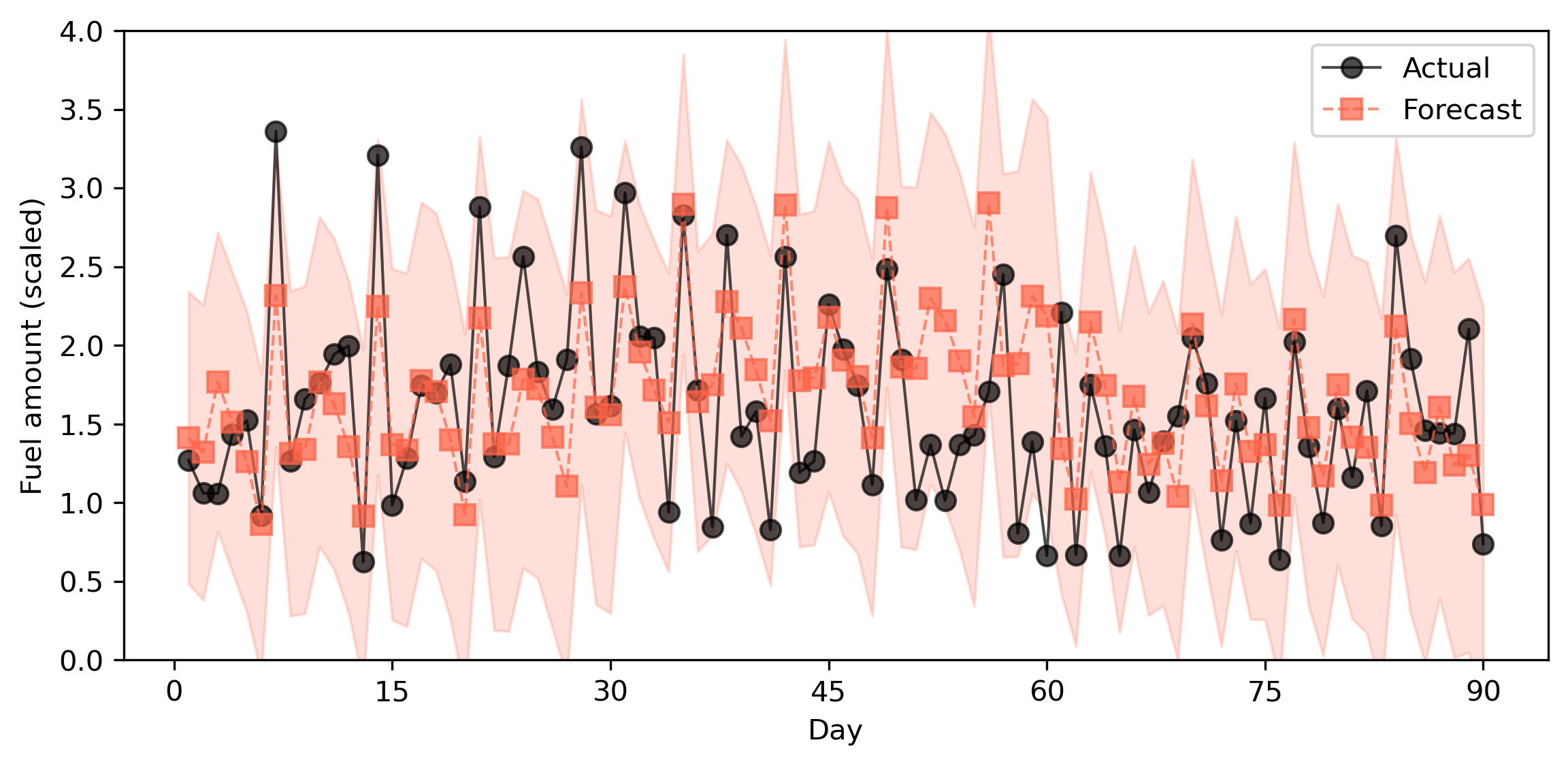}
\end{figure}

This case study is the only one for which the performance of the model previously used by the company has been provided, in terms of SMAPE. The accuracy of the expertise-based model used by the company had a performance of 17.5\%, which is much higher than the performance obtained by all the data-driven models. As we can see in Table \ref{tab:cph_scores}, the SARIMAX $(2,1,2)\times(1,0,3,7)$ model now emerges as the most effective forecasting model. In this case study, all the multivariate models surpass their univariate counterparts, underscoring the advantages gained from incorporating additional variables when predicting the overall volumes. By considering additional information related to the number and magnitude of customers operating their flights, these models showcase improved predictive performance. The test set predictions, displayed in Figure \ref{img:cph sarima}, demonstrate the SARIMAX model's ability to accurately capture the underlying time series patterns. The model exhibits remarkable reliability, as indicated by a low SMAPE value of 6.35\%. This implies that the predicted values closely align with the actual observed values, reinforcing the robust forecasting capabilities of the model. The successful performance of the SARIMAX model in predicting the fuel volumes at CPH suggests that the inclusion of exogenous variables provides valuable insights and contributes to enhanced forecasting accuracy. By leveraging the additional information about the number and characteristics of customers' flight operations, the model effectively captures the complexities of fuel demand at the airport. However, the univariate models also achieved a high degree of accuracy, offering improvements over the baseline established by the company.

\begin{table}
\centering
\begin{tabular}{@{}lccc@{}}
\toprule
\textbf{Model} & \textbf{RMSE} & \textbf{MAE} & \textbf{SMAPE} \\ 
\midrule
SARIMA  & 0.580 & 0.438 & 6.729 \\ 
SARIMAX  & \textbf{0.516} & \textbf{0.399} & \textbf{6.354} \\
Prophet univariate & 0.624 & 0.497 & 7.770 \\
Prophet multivariate & 0.616 & 0.487 & 7.596 \\
LSTM univariate  & 0.582 & 0.465 & 7.671 \\ 
LSTM multivariate  & 0.552 & 0.444 & 7.300 \\ 
Hybrid (SARIMAX-LSTM)  & 0.543 & 0.418 & 6.612 \\
\bottomrule
\end{tabular}
\caption{Test set prediction performance of the different methods on the third case study.}
\label{tab:cph_scores}
\centering
\end{table}

\section{Conclusion}
\label{sec:end}

The case studies presented in this paper demonstrate the effectiveness of data-driven approaches in forecasting fuel demand in the aviation industry, both at the airport and airline levels. The SARIMA and LSTM models consistently delivered competitive performance, highlighting their reliability in capturing the underlying patterns and fluctuations in fuel demand. The quality and availability of exogenous variables, such as flight schedules and customer information, play a crucial role in the modeling process. In cases where the data quality of these variables is not high, simpler models that do not incorporate exogenous variables are shown to provide more accurate predictions. However, when carefully selected and reliable exogenous variables are available, incorporating them into the forecasting models can yield additional benefits. The inclusion of these variables, as demonstrated in the third case study, improves predictive accuracy and provides valuable insights into future activity. Future research could also explore active and unsupervised learning paradigms to further improve data efficiency and model adaptability in operational contexts where acquire labeled data might be costly or time consuming \cite{cacciarelli2022novel,cacciarelli2023hidden,cacciarelli2025active,cacciarelli2024active}. Overall, the results of the case studies showcase the potential of data-driven approaches to enhance forecasting accuracy and support decision-making in the aviation industry. By leveraging machine learning models and analyzing historical data, fuel providers can gain a better understanding of future jet fuel demand and take informed sourcing decisions. This, in turn, enables more efficient resource allocation, accurate budgeting and planning, and ultimately leads to cost savings and increased profitability.

\typeout{}
\bibliography{sn-bibliography}

\end{document}